\documentclass[sigconf]{acmart}

\renewcommand\footnotetextcopyrightpermission[1]{} 

\renewcommand\footnotetextcopyrightpermission[1]{} 
\setcopyright{none}

\AtBeginDocument{%
  \providecommand\BibTeX{{%
    \normalfont B\kern-0.5em{\scshape i\kern-0.25em b}\kern-0.8em\TeX}}}

\copyrightyear{}
\acmYear{}
\acmDOI{}

\acmConference[]{}
\acmBooktitle{}
\acmPrice{}
\acmISBN{}

\usepackage{array}
\usepackage{amsmath}
\usepackage{algorithm}
\usepackage{comment}
\usepackage{mathtools}
\usepackage{booktabs}
\usepackage{multirow}
\usepackage{float}

\begin{document}

\title{Hierarchical Deep Reinforcement Learning Approach for Multi-Objective Scheduling With  Varying Queue Sizes}

\author{Yoni Birman, Ziv Ido, Gilad Katz and Asaf Shabtai}
\affiliation{%
  \institution{Department of Software and Information Systems Engineering \\ Ben-Gurion University of the Negev}
}

\begin{abstract}

Multi-objective task scheduling (MOTS) is the task scheduling while optimizing multiple and possibly contradicting constraints. A challenging extension of this problem occurs when every individual task is a multi-objective optimization problem by itself. 
While deep reinforcement learning (DRL) has been successfully applied to complex sequential problems, its application to the MOTS domain has been stymied by two challenges.
The first challenge is the inability of the DRL algorithm to ensure that every item is processed identically regardless of its position in the queue. 
The second challenge is the need to manage large queues, which results in large neural architectures and long training times.
In this study we present \MethodNameShort{}, a robust, modular and near-optimal DRL-based approach for multi-objective task scheduling. 
\MethodNameShort{} applies a hierarchical approach to the MOTS problem by creating one neural network for the processing of individual tasks and another for the scheduling of the overall queue. 
In addition to being smaller and with shorted training times, the resulting architecture ensures that an item is processed in the same manner regardless of its position in the queue. 
Additionally, we present a novel approach for efficiently applying DRL-based solutions on very large queues, and demonstrate how we effectively scale \MethodNameShort{} to process queue sizes that are larger by orders of magnitude than those on which it was trained. 
Extensive evaluation on multiple queue sizes show that \MethodNameShort{} outperforms multiple well-known baselines by a large margin ($\geq22\%$).
\end{abstract}


\renewcommand{\shortauthors}{}

\newcommand{\gilad}[1]{{\textcolor{blue}{[Gilad: #1]}}}
\newcommand{\asaf}[1]{{\textcolor{red}{[Asaf: #1]}}}
\newcommand{\yoni}[1]{{\textcolor{orange}{[Yoni: #1]}}}
\newcommand{\ziv}[1]{{\textcolor{cyan}{[Ziv: #1]}}}

\newcommand{\MethodNameShort}{MERLIN}

\newcommand{\argmax}[1]{\underset{#1}{\operatorname{\bf{arg}}\,\operatorname{max}}\;}

\DeclarePairedDelimiter\ceil{\lceil}{\rceil}

\newcolumntype{x}[1]{>{\centering\arraybackslash\hspace{0pt}}p{#1}}


\maketitle

\section{\label{sec:introduction}Introduction}

Scheduling algorithms plan the allocation of resources to tasks over a given time in order to optimize one or more evaluation metrics (e.g., throughput, average waiting-time in queue). 
The resources addressed by task scheduling can take on multiple forms: memory and CPU in a computing environment, machines in a workshop, runways at an airport, etc. 
Scheduling is crucial in multiple domains, including the manufacturing and service industries~\cite{hall1997scheduling}, medicine~\cite{may2011surgical}, and even malware detection~\cite{staniford2015systems}.
While simple scheduling tasks can be easily solved using existing heuristic  approaches~\cite{zhao1989performance}, multi-objective task scheduling (MOTS) problems are more challenging. 
The added difficulty stems not only from the need to balance multiple goals, but also from the need to sometimes reconcile \textit{contradictory metrics}.

In recent years, deep reinforcement learning (DRL)-based solutions have emerged as a promising alternative to existing heuristic scheduling solutions, often achieving state-of-the-art results~\cite{mao2016resource}. 
DRL algorithms have some significant strengths compared to other types of scheduling methods, particularly in cases that involve uncertainty~\cite{rl-uncertianty}.
First and foremost, they enable the formulation of sophisticated strategies, including those in which they make short-term sacrifices in order to reap larger rewards later on. 
Secondly, DRL algorithms are capable at efficiently exploring large state and action spaces, thus enabling them to develop novel and effective policies for complex scenarios. 
Thirdly, if the reward function (i.e., rewards and punishments for various actions and outcomes) is defined correctly, the DRL algorithm will very likely develop a strategy that achieves the desired goals (see~\cite{birman2019aspire} for a good example of how changing the reward function changes the algorithm's behavior).

While highly effective, DRL-based algorithms also have two significant shortcomings when applied to MOTS problems.
The first shortcoming of DRL-based approaches is the need to integrate multiple and often conflicting objectives into a single reward function. This is particularly the case when the processing of each task is fixed but instead constitutes a multi-objective optimization problem of its own. Such a function will need to address the multi-optimization problem both for the individual tasks and the entire queue, resulting in a complex and difficult-to-define expression. 
Moreover, each optimization goal is likely to have its own reward frequency and scale, thus contributing to the difficulty of defining the reward function. 
The use-case presented in this study (see Section~\ref{sec:usecase}) provides an excellent example of such a scenario: our goal is both to screen a set of files for malware, while minimizing the average processing time of each file. 
This challenge is further complicated by the fact that the analysis of each file is also a multi-objective optimization problem, with multiple detectors that can be used, each with its own capabilities and resource usage.

The second challenge associated with applying DRL-based solution to MOTS problems is that there no easy way of ensuring that samples are processed identically regardless of the state of the queue. This is the case because of the need to define a single reward function that models all constraints and priorities; integrating bounds and constraints into complex function is far from trivial.\newline
Consider a DRL-based system tasked with detecting malware in a queue of incoming files: the two (conflicting) goals set for the system are high detection rates (accuracy) and low average processing times.
If such a system becomes backlogged over time (i.e., average processing times rising quickly), the DRL-agent may begin conducting less extensive analysis of files (i.e., compromising detection rates) in order to reduce processing times. 
A scenario where an item is processed differently based on the current state of the queue is unacceptable in multiple domains, including medical testing and safety maintenance checks.

A more general shortcoming of applying DRL to scheduling problems -- one that is shared by all artificial neural network (ANN) architectures as well as other types of machine learning (ML) algorithms -- is that the size of the input used to train the model must be fixed. 
This requirement means that scheduling policies developed by such algorithms are unable to effectively operate on queues larger than the ones on which they were trained. 
This limitation can possibly lead to significantly sub-optimal solutions, as the algorithm can only process a part of the queue at each given time.
To (partially) address this problem, DRL-based scheduling algorithms are often trained on large queue sizes. 
While this course of action improves the algorithm's performance, it also requires considerably larger architectures, training data, and training time (see our results in Section~\ref{subsubsec:runningTimes}).

In this study we present \MethodNameShort{}, a hierarchical DRL-based scheduling approach for multiple objective scheduling.
Our proposed approach addresses the abovementioned shortcomings of other DRL-based approaches by applying a two-part solution. 
The first part of our proposed solution is creating a \textit{hierarchy of DRL agents}. 
Instead of attempting to solve the MOTS problem in its entirety using a single ANN architecture, we divide the problem into two parts: \textit{a)} devising a policy for the processing of individual items in the queue, and; \textit{b)} devising a policy for managing the queue. 
By using this approach we are able to accurately define the policy for the processing of specific items while also \textit{ensuring that each item is processed in an identical manner regardless of its position in the queue}. 
Additionally, the use of a modular solution results in smaller architectures that are easier to train than a single large architecture.

The second part of our proposed solution is a novel approach for enabling DRL-based scheduling solutions to manage queues larger than those on which they were trained. 
Our proposed approach is easy to implement and compatible with most DRL-based approaches. 
Moreover, our proposed solution enables the efficient processing of \textit{dynamic queue sizes}, i.e., queues where new items are stochastically added over time.  
To the best of our knowledge, the managing of dynamic queues was never addressed previously.

We evaluate \MethodNameShort{} on a large-scale malware detection dataset (presented in~\cite{birman2019aspire}). 
Our goal is to enable cost-effective analysis of the dataset's files: maintaining high detection rates while spending as little time as possible analyzing each file. 
The domain is challenging as it requires balancing both detection accuracy and the processing time of the entire queue. 
Our evaluation results shows that \MethodNameShort{} outperform multiple practical and realistic baselines used to evaluate our use-case by at-least 22\% across multiple queue sizes.

Our contributions in this study are: \textit{(1)} we propose a multi-objective scheduling framework that utilizes a multi-tier DRL solution. 
This approach simplifies the training process while enabling us to define various constraints; \textit{(2)} to the best of our knowledge, we are the first to present a DRL-based solution for scheduling \textit{with no prior data} both on the processed item and stochastic processing times, i.e., high uncertainty; \textit{(3)} we propose a hierarchical modeling approach that enables the processing of varying queue sizes without the need of retraining the algorithm; and \textit{(4)} we present an evaluation on a large real-world use-case (malware detection) that demonstrates the effectiveness and usefulness of our proposed scheduling framework.

\section{BACKGROUND: REINFORCEMENT LEARNING}

Reinforcement learning (RL) is a learning technique used for sequential decision making, often when only partial information is available or when the solution space is large. 
RL is highly adept at intelligently exploring various strategies in a highly efficient manner.
Because of their ability to successfully operate in complex domains, especially when coupled with deep learning, RL has been applied in domains such as robotics, control problems~\cite{schulman2015trust}, genetic algorithms~\cite{such2017deep}, complex games~\cite{silver2017mastering} and scheduling~\cite{mao2016resource}.

The RL problem setting consists of an \textit{environment} and an \textit{agent}. 
The agent takes \textit{actions} that affect the environment and change its \textit{state}. 
Each action (or sequence of actions) incurs a \textit{reward} that provides feedback to the agent on the quality of its decisions. 
Agents can optimize their behavior by interacting with the environment and devising a \textit{policy} that will yield maximal rewards overall. At every time-step $t$, the agent selects an action $a_t$ from the action space $A=\{a_1, a_2,...,a_k\}$ that modifies the state of the environment and incurs a reward $r_t$ (positive or negative).
The goal of the agent is to maximizes future accumulated reward $R_t= \sum_{t}^{T} r_t$ where $T$ is the index of the final time-step.

A common approach for selecting the action to be taken at each state is the action-value function $Q(s,a)$~\cite{p103}, also known as the $Q$-function. 
The function approximates the expected returns should we take action $a$ at state $s$. 
While the methods are varied, RL algorithms which use $Q$-functions aim to discover (or closely approximate) the optimal action-value function $Q^*$ which is defined as 
$Q^*(s,a)=\max_\pi E[R_t|s_t=s, a_t=a, \pi]$ where $\pi$ is the policy mapping states to actions~\cite{p103}. 
Since estimating $Q$ for every possible state-action combination is highly impractical~\cite{p98}, it is common to use an approximator $Q(s,a;\theta)\approx Q^*(s,a)$ where $\theta$ represents the parameters of the approximator. 
Deep reinforcement learning (DRL) algorithm perform this approximation using neural nets, with $\theta$ being the parameters of the network.

\section{\label{sec:method}THE PROPOSED METHOD}
\subsection{Motivation}
While DRL has proven very effective in optimizing a single objective~\cite{mao2019learning}, to date no study has successfully applied this approach to multi-objective scheduling (although multi-resource problems have been addressed~\cite{mao2016resource}).
This is likely due to the difficulty of balancing multiple, and often contradicting, objectives in a single reward function. 
When multiple goals affect the reward function, it is more difficult to isolate the effect each single action has on each of the objectives. 
Moreover, there is an inherent difficulty in integrating objectives that may have different value scales and distributions and are provided at different intervals.

Our proposed approach partitions the original problem into separate ``sub-problems'', each solved with its own DRL-agent. 
Such partitioning simplifies each individual problem, enables modularity, and reduces the complexity of the overall optimization process. 
It is important to note that each sub-problem does not have to address all the goals of the original problems, but instead can solve a subset of the said goals.
This can be done by defining additional (intermediary) goals in order to facilitate the desired outcome for each sub-problem. 
Such a use of an intermediary goal is presented in our own use-case in Section~\ref{sec:usecase}.

We define our approach as \textit{modular} since it enables us to easily replace each of the individual DRL-agents used to solve the sub-problems. 
This trait is very useful from a practical standpoint since it enables us to re-calibrate our model (e.g., change some of our priorities---greater accuracy and lower throughput) without re-training all of its components.
Our approach was inspired by the modular NN approach \cite{kimoto1990stock}, which applies this idea in the field of robotics.

\subsection{Problem Formulation}
In order to simplify our representation, we present a problem formulation for a two-tier hierarchical model. 
The proposed representation can easily be expanded to include additional tiers.

Let $Q = \{q_1,...,q_n\}$ be a queue of $n$ items. 
Let ${R_j^i}$ be the \textit{internal reward function}, which defines the reward (positive or negative) obtained when processing item $q_j$. 
Let $R_Q^o$ be the \textit{outer reward function}, which defines the reward for processing the entire queue. 
We begin by optimizing the loss function of the internal agent, thus setting the policy of the internal agent defined by $\theta_i$:
\begin{center}
$\argmax{\theta_i} \sum_{n=1}^{|Q|} \mathcal{L}_i(Q_n,R^i,\theta_i)$    
\end{center}
where $Q_n$ is the n\textsuperscript{th} item in $Q$ and $\mathcal{L}_i$ is the loss function of the internal agent.

Once we set the policy of the internal agent $\theta_i$ we can define the policy of the outer agent. 
Our goal is to minimize the loss function and optimize the policy of the outer agent $\theta_o$.

\begin{center}
$\argmax{\theta_o} \mathcal{L}_o(Q,R^o,\theta_o|\theta_i)$    
\end{center}
where $\mathcal{L}_o$ is the loss function of the outer agent.\\

\noindent \textbf{Motivation.} Our proposed approach for addressing this challenge consists of a \textit{modular and hierarchical} DRL architecture. 
We begin by setting all of the problem domains' constraints and priorities, both for each individual item in the queue (e.g., desired detection rate for defects) and for the queue overall (e.g., average processing time).
We define these priorities by $R^i$ and $R^o$, respectively. 
We then train one DRL agent -- the ``internal agent'' -- to create the \textit{internal policy} $p^i$ that optimally addresses $R^i$. 
Finally, we ``freeze'' $p^i$ and use a second DRL agent -- the ``outer agent'' -- to train the \textit{outer policy} $p^o$, whose goal is to optimize $R_Q^o$ by scheduling the order by which $p^i$ is applied on the items in $Q$. 
The inputs for $p^o$ are both the current state of the queue and the outputs of the internal-agent.

It is important to point out that we first set the internal policy and only train attemp to optimize the outer policy. This appraoch ensures that \textit{each item in the queue is processed in the same manner regardless of its position in the queue}. This trait, which cannot be guaranteed in DRL architectures that use a single reward function for the entire problem, ensures consistency in performance and equal treatment of all items in the queue regatdless of their position in the queue. Guarantees such as the ones our approach provide are critical in many fields, including medical testing and airplane maintenance.

\subsection{System Architecture}
\label{subsec:sysarch}
\noindent \textbf{Modular architecture.}

\begin{figure}[h]
  \centering
  \includegraphics[scale = 0.24]{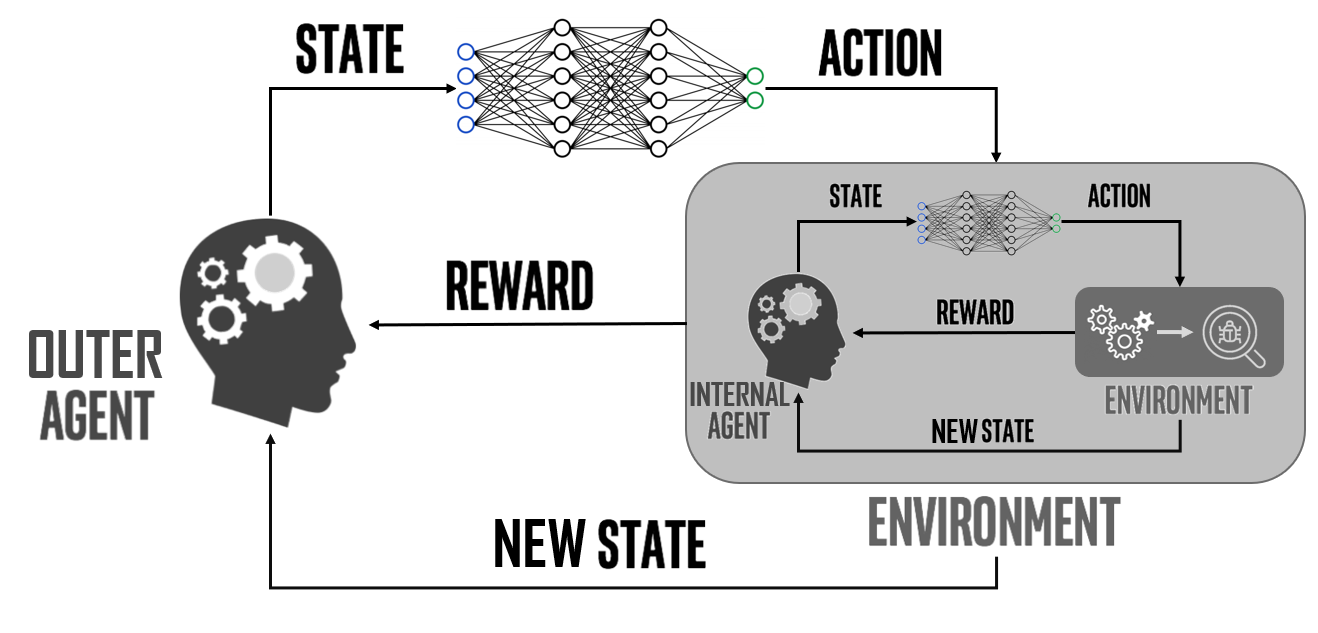}
  \caption{Illustrating our two-tier, modular DRL-based scheduling algorithm.
   The internal agent uses an optimal policy defining how to process a single item in the queue. The outer agent's goal is to serve as the scheduling mechanism for the queue.}
  \label{fig:doublearch}
\end{figure}

\noindent Our proposed solution architecture consists of an outer agent whose goal is to schedule the processing of the various items in the queue, and an internal agent whose goal is to determine the manner by which each individual item is processed (see Figure~\ref{fig:doublearch}). 
As mentioned before, each agent is trained separately: the internal agent is trained first until it converges. 
Then, the outer agent is trained by interacting with the internal agent, i.e., exploring various scheduling strategies that involve the fully-trained internal model. 
The internal agent is ``frozen'' while the outer agent is trained, ensuring both modularity and that the desired performance of the internal agent---which was defined during its training---is maintained.

\noindent \textbf{Multiple objectives.}

\begin{figure}[h]
    \centering
    \includegraphics[scale = 0.3]{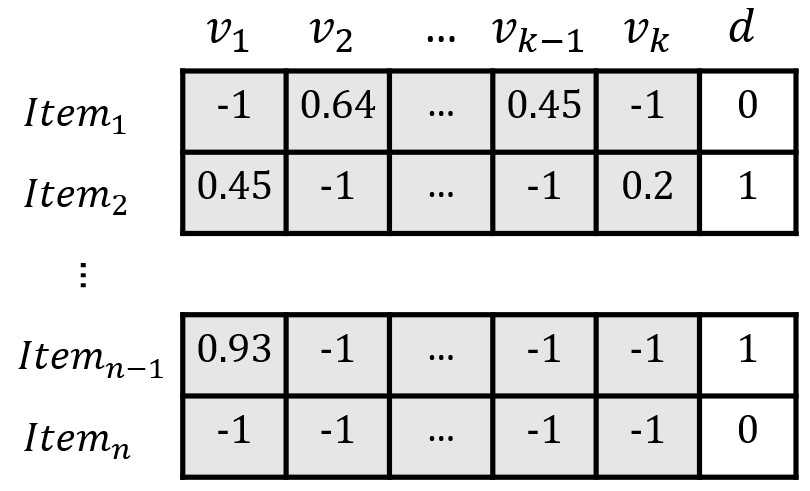}
    \caption{An example of a state matrix with $n$ items.
    The inner item representation consists out of $k$ values.
    $d$ indicates whether the processing of the item has ended. Note that $v_k$ is the $k$\textsuperscript{th} value in the inner item representation.}
    \label{fig:state_doublel}
\end{figure}

\noindent The roles of the two agents are very different: The goal of the internal agent $p^i$ is to create an \textit{optimal policy for a single item in the queue}. 
As a result, the state of the internal agent $s^i$ represents the \textit{current state} of a single item $i$. 
The goal of the outer agent $p^o$ is to serve as the \textit{scheduling mechanism for the queue}.
Therefore, the state representation of outer agent $s^o$ is a concatenation of all current item representation $S^o=\{s^i_1, s^i_2,...,s^i_n \}$, with the addition (to each item state $s^i_j$) of a single value $d$ indicating whether the processing of the item has ended. 
An example of the state representation of the outer agent is presented in Figure~\ref{fig:state_doublel}, where each row represents an item, and each column represents an action that can be taken by the internal agent on that item. 
Non-negative cell values are the outputs of executed actions.

\noindent \textbf{Actions selection.}

\noindent At every time step, the outer agent can select the item that will be sent to the internal agent for processing. 
As a result, the size of the outer agent's action space is equal to the size of the queue. 
The size of the internal agent's is determined by the number of processes it can apply on an item (columns in Figure~\ref{fig:state_doublel}), with an additional action for final classification.

It is important to note that outer agent's scheduling is \textit{preemptive}. 
This means that once an item has been submitted to $p^i$, the internal agent \textit{only performs a single action} rather than analyze the item until completion. 
$p^o$ can then choose to send a different item to $p^i$, leaving the first item to be processed at a later time. 
The rationale of using this approach is simple: since each action taken by $p^i$ reveals additional information the item, $p^o$ has a chance to weigh the benefit of continuing to process the current item against processing another. 

The internal agent $p^i$ is also preemptive in the sense that it can issue a final decision about the analyzed item, without having to run all possible test/processes on it. This setting, initially proposed in \cite{birman2019aspire}, enables the internal agent to strike the desired balance between performance (e.g., classification accuracy) and the resources allocated to achieving it.\\

\noindent \textbf{Operating under high levels of uncertainty.} 

\noindent We argue that due to its being a DRL-based method, \MethodNameShort{} has an important advantage over existing solutions when dealing with high degrees of uncertainty. The uncertainty presents itself in two ways in our use-case: first, the internal agent has no way of knowing in advance the output provided by each detector. Additionally, the runtime of each detector varies from file to file, thus adding another level of complexity to the malware detection process. Secondly, the outer agent has to contend with uncertainty regarding the actions and running time of the internal agent as it selects the next files to be processed by the latter.

Unlike other commonly-used approaches~(See Section \ref{subsec:SchedulingBaselines}), \MethodNameShort{} requires \textit{no preliminary information} on the processed items---file size, file type, bounds on running time, etc.---and adapts its policy by interacting with the items over time.
The ability to operate under high uncertainty is shared both by $p^i$ and $p^o$. 
$p^i$ interacts with individual items and devises its own policy for processing them.
$p^o$ interacts both with items of the queue and with $p^i$, without any prior knowledge on either.

\subsection{Adapting to Changes in Queue Length}
\label{subsec:dynamic_handling}
One significant shortcoming of DRL-based solutions to queue management is the network's inability to adapt to changes in the state or action space \cite{schulman2015trust}. 
More specifically, we refer to the fact that the input of the network, and consequently its number of actions, must be of a fixed size. 
This inflexibility leads to two types of problems. First, this could easily lead to sub-optimal solutions, with easy-to-process items having to wait until the first $X$ items are done. 
An example of this scenario is presented in \cite{mao2016resource}: since the value of $X$ was 100, the $X+1$ item will not be considered until the first $X$ items are completed.
Secondly, this inability to infer the learnt logic to larger queue sizes forces practitioners to train their DRL-agents on relatively large state representations, a fact that leads to longer running times and difficulties for the deep network to reach convergence.

To address the challenges described above, we propose a novel \textit{hierarchical approach for dynamic queue size scheduling}. 
Given a queue $Q$, we partition it into fixed size subsets of size $n$, where $n$ is also the number of queue items the outer agent $p^o$ is configured to receive as input. 
This partitioning results in $\ceil{\frac{|Q|}{n}}$ sub-queues.
To ensure that all sub-queues are exactly of size $n$, we use padding when needed. 
Our padding consists of items that are flagged as ``completed'' (i.e., their processing is already complete), which effectively ensures that our fully-trained DRL-agent will ignore them.

\begin{figure}[ht]
    \centering
    \includegraphics[scale = 0.275]{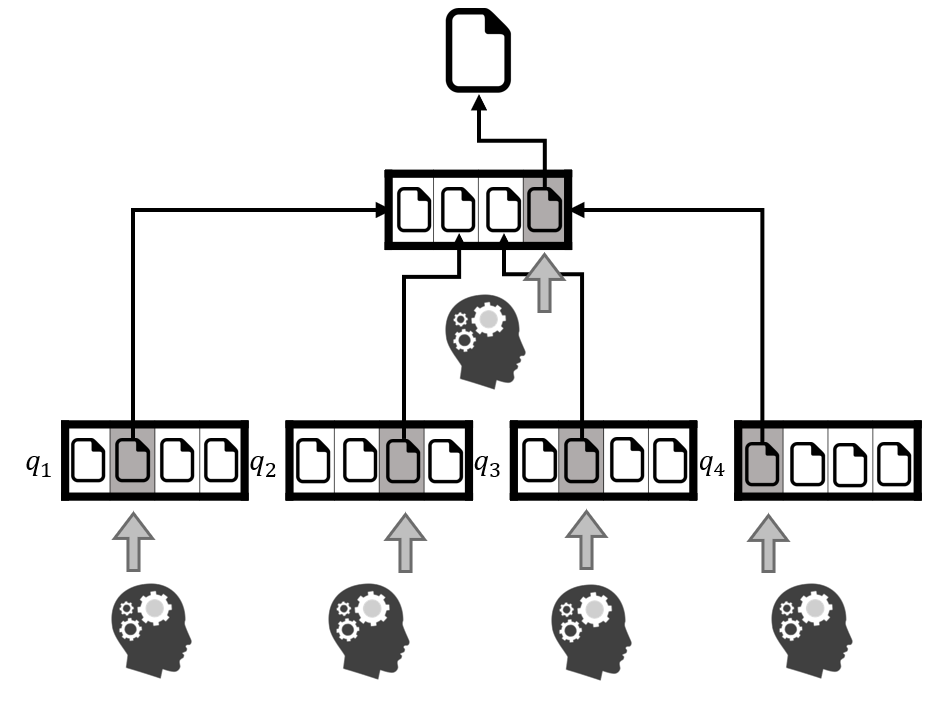}
    \caption{An example of two-stage hierarchical state-action reduction process, with trained DRL algorithm on queue with length of four.}
    \label{fig:hierarchial_step}
\end{figure}

Once the partitioning into sub-queues is complete, we apply the outer agent $p^o$ on each sub-queue. 
This results in the creation of a selected set of queue-items whose size we denote as $|Q^`|$. 
Next we check to see whether $Q^` \leq n$. 
If that is the case, $Q^`$ is provided as input to $p^o$ and the scheduling process continues as described in Section \ref{subsec:sysarch}. 
Otherwise, if $Q^`> n$, we once again partition the current set into $\ceil{\frac{|Q^`|}{n}}$, and continue to do so iteratively until we reach an item set of size $n$. 
An illustration of the proposed process is presented in Figure \ref{fig:hierarchial_step}.

The proposed hierarchical approach has two significant advantages. 
First, it enables the use of a DRL algorithms with fixed-size input representation for processing queues of practically any size, thus removing one of the main limitations to applying DRL to queue management. 
Moreover, the item processing is done in a way that looks at every item on the queue---no item is ignored. 
Secondly, our hierarchical approach makes it possible to train networks with smaller input sizes to process large queues. 
As a result, we can train smaller networks with less parameters, thus leading to faster convergence and the need for fewer computing resources. The results of our evaluation in Section~\ref{subsec:results} support our claim.

\section{USE-CASE: MALWARE DETECTION}
\label{sec:usecase}

Our use-case is based on the study presented in~\cite{birman2019aspire}, where a DRL-based framework was used to perform cost-aware analysis of malware files.
The underlying insight of the said study is that while organizations often deploy an ensemble of detectors to ensure high detection rates, in many cases a subset of the available detectors can produce the correct classification using far fewer computing resources and shorter execution times. 
The authors of~\cite{birman2019aspire} created a reward function that factors in both correctness of the classification and the time needed to reach the decision, and show that their approach can significantly reduce the time needed to classify a file (by $\sim$80\%) while only marginally harming detection accuracy.

It is important to note that the solution presented in \cite{birman2019aspire} is designed for the cost-effective classification of \textit{individual files} and not to the management of queues. For this reason, we use the DRL agent developed in \cite{birman2019aspire} as the internal agent in this use case, and then train the outer agent ourselves. For a comprehensive overview of the architecture, we refer the reader to the original paper. 
The remainder of this section provides a short overview on the internal and outer agents' architecture. \\

\subsection{The Internal Agent} 

The goal of the internal DRL-agent is to create a cost-effective policy for the analysis of files for possible malware.
To achieve this goal, the agent performs the following steps: \textit{1)} send a file to one detector; \textit{2)} receive the classification output of the detector (a value in the range [0,1]); \textit{3)} based on the available information, determine whether to provide a final classification to the file, thus terminating the process, or query an additional detector(s). If the latter option is chosen, all steps are repeated.

The reward functions evaluated in \cite{birman2019aspire} are presented in Table \ref{tab:experiments_2}. All functions define the cost of making a mistake (i.e., false-positive of false-negative) as a function of the time spent classifying the file. The logic behind this approach is simple yet novel: discourage the DRL-agent from querying detectors that are unlikely to provide useful information (i.e., increase chanced of being correct), as they will only lead to more ``pain'' in case of a mistake. The reward for correct classifications is either fixed, or a function of the time spent. The former option encourages the algorithm to be more cost-oriented, resulting in shorter processing times per file. The latter approach yields superior performance but saves only a little amount of computing resources. In our study we chose to use the reward function of experiment \#3, which offers (in our view) the best cost/effective solution (savings of about 80\% in running time while reducing performance by only 0.5\%). This is the policy used in all our experiments throughout this study.

The state space of the internal agent is represented by a vector containing a single entry for each of the available malware detectors. The value of each cell is either -1 (minus one)---meaning that the detector was not yet queried---or containing the output of the detector, a value in the range $[0,1]$. 

The action space of the internal agent is similarly simple: it contains one action for each detector, and choosing this action will query the corresponding detector to classify the file. Additionally, there are two more actions: \textit{1)} classify file as benign, and; \textit{2)} classify file as malware. Choosing one of the two latter actions terminates the analysis of the file.

\begin{table}[ht]
  \caption{The five reward setups evaluated in \cite{birman2019aspire}. In the experiments presented in this study, we use the configuration of experimetn \#3.}
  \label{tab:experiments_2}
  \setlength\tabcolsep{3.9pt}
  \small
  \begin{tabular}{c|cccc|cc}
    \toprule
    Exp.& \multicolumn{4}{c|}{Reward Setup} & Accuracy & Mean\\
     \# & TP & TN & FP & FN & (\%) & Time (sec)\\
    \midrule
    1 & C'(t)  & C'(t)  & -C'(t)   & -C'(t)   & 96.867 & 48.61 \\
    2 & C'(t)  & C'(t)  & -10C'(t) & -10C'(t) & 96.801 & 48.37 \\
    3 & 1      & 1      & -C'(t)   & -C'(t)   & 96.212 & 10.53 \\
    4 & 10     & 10     & -C'(t)   & -C'(t)   & 95.424 & 3.68  \\
    5 & 100    & 100    & -C'(t)   & -C'(t)   & 91.220 & 0.73  \\
  \bottomrule
\end{tabular}
\end{table}

\subsection{The Outer Agent}
The goal of the outer agent is to schedule the processing of the analyzed files by the internal agent. The goal of the outer agent is to minimize the average time a file spends in the queue while waiting to be classified. A detailed description of the evaluation metric is provided in Section \ref{experimentalSetup}.

The state space of the outer agent is modeled using a matrix like the one presented in Figure~\ref{fig:state_doublel}. 
Each row in the matrix represents a single file in the queue, and it consists of the file's state, as it is represented by the internal agent. Simply put, the state space of the outer agents is a concatenation of the internal agent's state representation for all files. The action space of the outer agent is equal to the initial number of files in the queue. Choosing action $A_i$ indicates that the $i^{th}$ file is sent to the internal agent for processing.

\noindent It is important to note that the outer agent interacts with the internal agent as a black-box model. The outer agent has no information regarding the internal agent's inner workings or decision-making process. The outer agent develops its own policy simply by interacting with internal agent and inferring on its own the optimal policy. This setting is both simple and more robust, as it enables modular training and the replacement of either the internal or outer agents.

\section{\label{sec:evaluation}Evaluation}
\subsection{Experimental Setting}
\label{experimentalSetup}

\begin{figure}[ht]
  \centering
  \includegraphics[scale = 0.3]{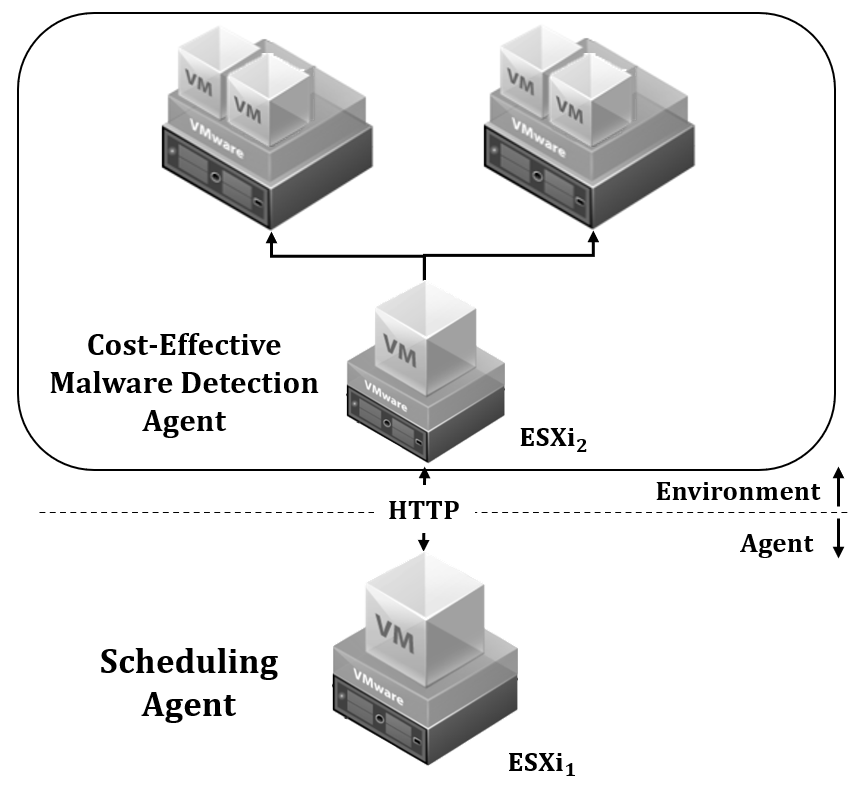}
  \caption{The experimental infrastructure's architecture.
  $ESXi_1$ hosts the outer agent and the PE files. 
  $ESXi_2$ hosts the internal agent.}
  \label{fig:infrastructure}
\end{figure}

\noindent \textbf{Hardware setting.} For our experiments, we used the VMware ESXi operating system for our servers, each with two processing units (CPU). 
The server had total of 32 cores, 512GB of RAM and 100TB of SSD disk space.
Figure~\ref{fig:infrastructure} provides a comprehensive overview of the infrastructure we used.
The outer agent process ran on virtual machine (VM) with the Ubuntu 18.04 LTS operation system. 
The virtual machine had 16 CPU cores, 16GB of RAM, and 10TB of SSD storage.
The agent uses a management service that allows both the training and execution of the DRL algorithm, using different tuning parameters.

The internal agent process ran, according to the specified specification provided in~\cite{birman2019aspire}, on three VMs with Ubuntu 18.04 LTS operation system. 
Each machine had 4 CPU, 16GB RAM configuration with additional 100GB of SSD storage.
Upon the arrival of files for analysis, the agent stores them in a logical queue at a dedicated storage space, which is also accessible to the internal agent.
Both agents use an external storage to all logging information in an indexing engine for search and analysis capabilities. 

\noindent \textbf{The dataset.} We evaluated \MethodNameShort{} on the dataset presented in \cite{birman2019aspire}, which consists of 25,000 executable files, half malicious and half benign. 
We obtained both the dataset and the reported running time of each detector for every file, which enabled us to train a DRL agent that accurately replicate the the results reported in \cite{birman2019aspire}. 
As explained in Section~\ref{sec:usecase}, we use this architecture as our internal agent.\\

\noindent \textbf{Training and the evaluation measure.} We trained our outer agent $p^o$ with the goal of optimizing the scheduling process of all files. 
The metric used both to train the outer agent and evaluate the overall performance of our approach was \textit{average job completion time}~\cite{hall1997scheduling}.  
Let $N$ be the number of files in the queue, $PT_{q_i}$ the total processing time of $q_i$ in the internal agent, and $WT_{q_i}$ the waiting time of $q_i$ in the queue. The completion time for item $q_i$ in queue is $C_{q_i} = PT_{q_i} + WT_{q_i}$. The \textit{average completion time} is calculated as shown in Equation~\ref{eq:completiontime}.
In essence, this metric is used to measure the average processing time for each file in the queue.

\begin{equation}
   \textit{average completion time} = \frac{{}\sum_{i=1}^{N} C_{q_i}}{N}
   \label{eq:completiontime}
\end{equation}

It is important to note that the experimental setting contains of a \textit{high degree of uncertainty}. 
Our approach uses no prior information about the analyzed files (not even their size, which is used by some of our baselines in Section \ref{subsec:SchedulingBaselines}). 
All available information on a given file is obtained solely through its processing (i.e., sending it to the internal agent).
Additionally, we trained \MethodNameShort{} \textit{on a queue-size of 10 items}. 
By doing so, we demonstrate our model's ability to easily scale for larger queues, sometimes larger by orders of magnitude, despite of the high uncertainty of the dataset.\\

\noindent \textbf{Hyperparameter settings.} We used the following settings throughout our evaluation. The train/test split was 90\%/10\%, with the same setting used for all experiments. The DRL architecture used was actor-critic with experience replay~\cite{p97}. The outer agent was trained for 34 epochs, which required 22 hours. The framework was implemented on OpenAI Gym, using python version 3.6. Both DRL agents -- $p^i$ and $p^o$ -- used an actor-critic architecture with a single hidden layer of size 20. 
The hidden and output layers used ReLU and Softmax functions respectively. 
We used a replay buffer of size 10, which was activated after 1000 episodes. 
We used a learning rate of $7e^{-5}$, an exponential decay rate of $0.99$ and a fuzz factor (epsilon) of $1e^{-2}$. 
We also used RMSprop~\cite{rmsprop} optimization. 
All experiments were trained until convergence. 
We used penalties to discourage the agent from taking illegal actions (i.e., selecting files that were already classified). 

\subsection{Scheduling Algorithms Baselines}
\label{subsec:SchedulingBaselines}
We compare \MethodNameShort{} both to ``naive'' solutions and to well-known scheduling algorithms. 
All of our chosen baselines are known to function well in high levels of uncertainty. 
Additionally, all baselines are able to seamlessly operate both on different queue-sizes and on dynamic queues where additional items arrive stochastically.

It is important to note that all baselines are ``competing'' against the outer agent $p^o$, i.e., they all select the order of the files to be sent to $p^i$.  
This was done for two reasons: First, these baselines are scheduling algorithms, and therefore cannot perform the classification process of individual files. 
Secondly, by using the same internal agent for all algorithms we ensure that the performance in terms of classification accuracy is uniform. We can therefore evaluate the various algorithms based on their running times. We wish to stress this point again, as it is crucial to understanding our evaluation: \textit{using different internal agents or allowing the internal agent to change its policy throughout the evaluation will lead to different detection rates and thus make the comparison between the algorithms impossible.}

Because of the high uncertainty of our problem definition (i.e., no prior information exists on the analyzed files), several commonly-used scheduling approaches could not be used in our experiments. 
In order to overcome this limitation, we define two groups of baselines: a ``realistic'' group in which the baseline algorithms have access to the same information as our approach, and an ``unrealistic'' group in which the baselines have access to additional data that is not available to \MethodNameShort{}. 
Our evaluation shows that \MethodNameShort{} outperforms both groups (except for the baseline which serves as an optimal lower bound).

\subsubsection{``Realistic'' Baselines}
This group consists of four baselines, all with access to the same information as \MethodNameShort{}. 
It should be noted that two baselines in this group (SFF and LFF) also use the sizes of the analyzed files. \MethodNameShort{} does not use this information, but since this information can be easily achieved we include these baselines in the current group.

\noindent \textbf{First Come First Serve (FCFS)}. A naive scheduling algorithm, that schedules tasks according to their initial position in the queue. 
In our use-case (see Section \ref{sec:usecase}), once a file reaches the top of the queue, it is processed by the internal agent until a classification decision is reached (i.e., ``malware'' or ``benign'').
    
\noindent \textbf{Smallest File First (SFF)}. A variant of the shortest job first (SJF) approach.
Assuming that a smaller file is likely to require less time to classify, the algorithm sorts the files in the queue based on their size, in an  ascending order. The files are then sequentially processes until completion. 
We have also tested an \textit{inverted version} of this scheduler: the longest file first (LFF) algorithm.
    
\noindent \textbf{Multi Level Feedback Queue (MLFQ)}. A priority queue-based algorithm that allocates items to multiple sub-queues based on their required resources. 
In our experiments we used three sub-queues that partitioned the items based on the running time of next detector assigned to them by the internal agent $p^i$ (i.e., the time of the next action to be performed on the file). 
Once the detector was applied on the file, $p^i$ determines (but doesn't execute) what is the next detector that needs to be used. 
Based on the running time of that detector, the file was then assigned to the appropriate sub-queue. In case where the next action was final classification, the item was removed from the queue. 

\subsubsection{``Unrealistic'' Baselines}    
Each baseline in this group has access to information that is either unavailable to \MethodNameShort{} (e.g., knowledge on general processing times distributions) or ``oracular'' (knowledge of specific running times in advance). For each baseline, we specify the specific information available to it.

\noindent \textbf{Shortest Expected Processing Time (SEPT)}. This baseline implements a stochastic scheduling approach. Since approaches of this type require knowledge about the distribution of the overall processing time of the population~\cite{smith1956various}, we extract this information from the training set prior to running the scheduling algorithm.

\noindent \textbf{Correlation Based Processing Time (CBPT)}. This baseline assumes that we have, in advance, the classification results (i.e., confidence score) of one of the malware detector for all files. Based on these scores, we sort the files in the queue \textit{according to their likelihood of being benign}. Since benign files usually require less analysis than malicious ones, since the internal agent usually processes them more quickly, this is a high-performing baseline. For this task we chose the detector with the highest Pearson correlation between its confidence scores and the true item labels.

It is important to note that we treat the confidence scores used for the item-ranking as prior knowledge, meaning that the internal agent may call the detector that produced the classifications as part of its analysis. Despite its relatively high performance (see Figure \ref{fig:eval_baseline10}), we consider this baseline as unrealistic since applying even a single detector in advance to \textit{all} the items in large queue sizes will lead to very poor performance due to the long time it would take to produce a classification.

\noindent \textbf{Shortest Processing Time (SPT)}. The files are ordered in an \textit{ascending order}, based on their total classification time. In other words, we have perfect information on the time needed to classify each file.  This baseline in guaranteed to achieve the top performance.

\noindent \textbf{Longest Processing Time (LPT)}. The files are ordered in an \textit{descending order}, based on their total classification time. This baseline is guaranteed to achieve the worst performance.

\subsection{\label{subsec:results}Experimental Results}

We conducted three types of experiments, each with an increasing degree of complexity. 
We began by evaluating \MethodNameShort{} on the queue size on which it was trained (10 items). 
Then, we evaluated our approach's ability to perform on larger queue sizes. 
Finally, we evaluated a dynamic queue where new items are being stochastically added.

\subsubsection{Experiment 1: Fixed Queue Size}

This evaluation was conducted on $|Q|=10$, which is also the input size of our DRL-agent. To ensure the validity of our results, we randomly sampled 10 files from our test set and provided it to all evaluated algorithms. This process was repeated 2,500 times, with the presented results being the average performance across all runs.

\begin{figure}[h]
    \centering
    \includegraphics[scale = 0.49]{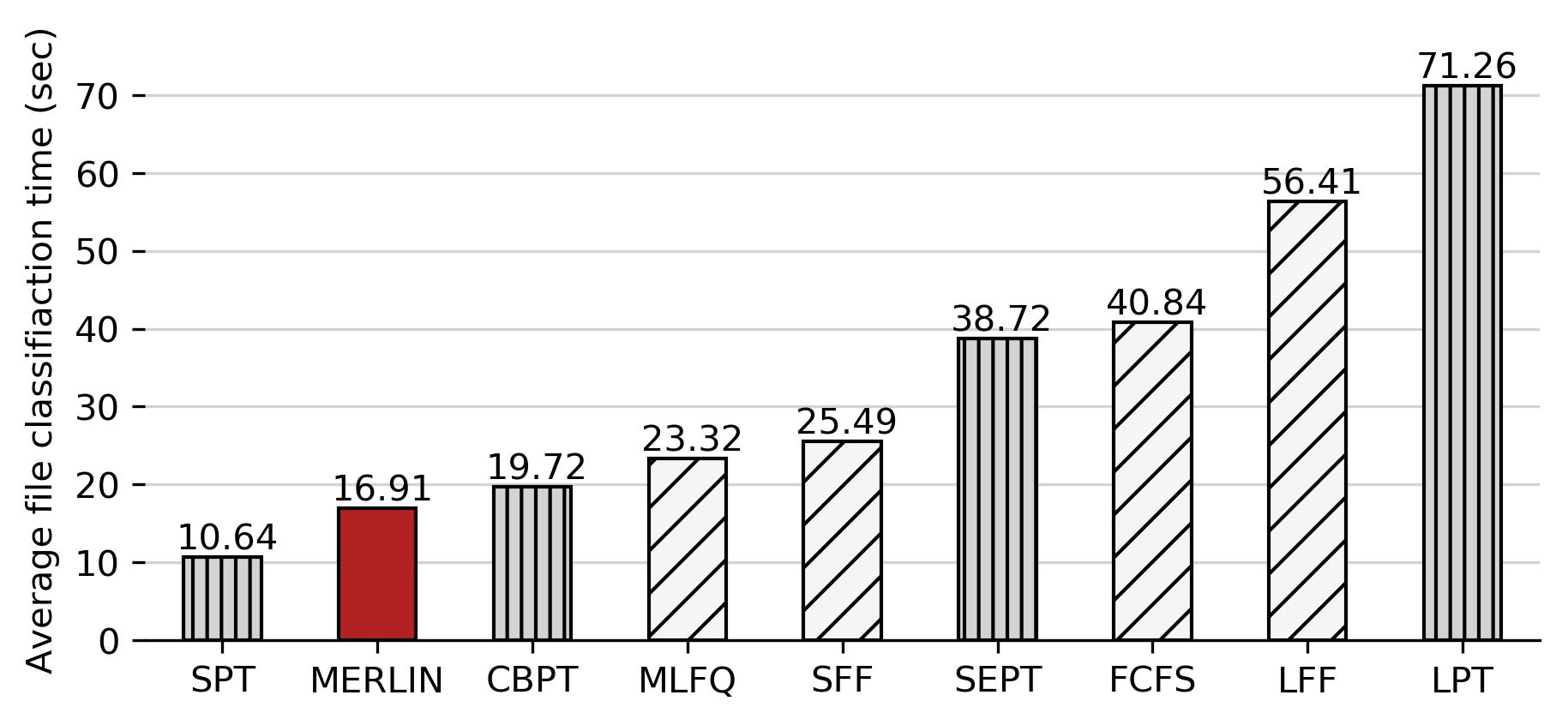}
    \caption{Average completion time comparison in a 10-items queue (diagonal stripe: "realistic" baselines; vertical stripe: "unrealistic" baselines).} 
    \label{fig:eval_baseline10}
\end{figure}

The results of this experiment, presented in Figure~\ref{fig:eval_baseline10}, clearly show that our approach outperforms all the evaluated baselines (except for the SPT, which is the optimal scenario). 
When compared to ``realistic'' baselines (diagonal stripe columns in Figure~\ref{fig:eval_baseline10}), the average job completion time is shorter by 27\%-71\%. 
When compared to ``unrealistic'' baselines (vertical stripe columns), the average job completion time is shorter by 14\%-57\% . 

Our analysis indicates that the reason for \MethodNameShort{}'s superior performance is its ability to better handle uncertainty. 
Since both the outer and internal agents need to address uncertainty (albeit at different aspects of the challenge), the outer agent's ability to infer the internal agent's policy and behavioral patterns enables it to create its own complementary policy. 
The interaction between the two DRL-agents is particularly evident in their dealings with difficult-to-classify files, which are files that the internal agent would not classify with applying multiple detectors. 
The outer agent identifies these time-consuming files early on (usually after the confidence score of the first detector is produced) and immediately pushed such files to the back of the queue in order to finish with the ``easier'' files first.

\subsubsection{Experiment 2: Large Queue Size}

Next we evaluate \MethodNameShort{}'s ability to perform well on varying queue sizes. We ran the same experimental setup as Experiment 1, but with queue sizes ranging from 10 to 100. 
For each queue size, we generated 1,250 random queues which were used by all algorithms for evaluation. 
It is important to note that the \MethodNameShort{} architecture used in \textit{all} experiments was trained in a queue size of 10.

\begin{table}[h]
    \centering
    \small
    \caption{The average completion time for the different algorithms for a single file over queue sizes ranging from 10 to 100.}
	\begin{tabular}{ccccccc}
		\toprule
		Queue & \MethodNameShort{}& CBPT& MLFQ & SFF& FCFS& LFF \\
		\midrule
        10 &  \textbf{16.91} &19.72& 23.32 & 25.49 & 40.84 & 56.41  \\
        20 & \textbf{28.74} &32.53& 37.14 & 45.26 & 78.28 & 111.14  \\
        30 &  \textbf{41.01} &45.31& 50.82 & 64.32 & 115.38 & 166.53  \\
        40 & \textbf{50.44} & 58.08&64.96 & 83.40 &  152.92 & 221.96  \\
        50 &  \textbf{62.55} & 70.89&79.15 & 103.35 & 189.05 & 276.65\\
        60 &  \textbf{74.22} & 83.89& 92.58 & 122.04 & 227.22 & 332.22\\
        70 &  \textbf{85.05} & 96.55&106.75 & 141.61 & 264.95 & 387.17 \\
        80 &  \textbf{95.44} & 109.29&120.64 & 160.80 & 302.00 & 442.48 \\
        90 &  \textbf{105.48} & 122.26&134.37 & 178.20 & 339.48 & 499.50 \\
        100 & \textbf{115.50} & 134.97&148.60 & 198.70 & 374.90 & 553.80  \\
		\bottomrule
	\end{tabular}
	\label{tab:eval_baseline_small}
\end{table}

\begin{figure}[h]
    \centering
    \includegraphics[scale = 0.5]{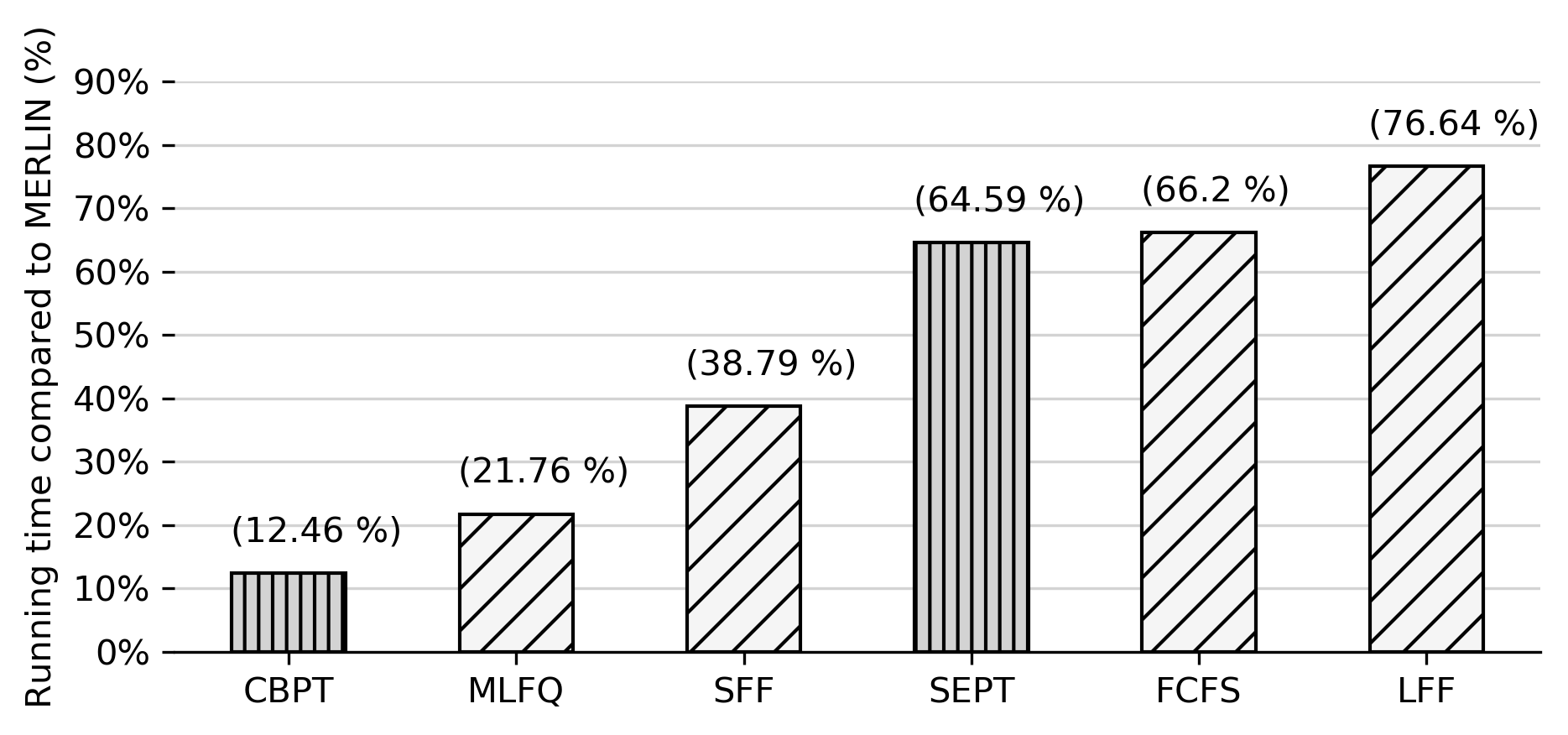}
    \caption{The relative performance of the baselines compared to \MethodNameShort{}. Higher percentage means longer running times compared to our method.}
    \label{fig:eval_im_summary}
\end{figure}

The results of our evaluation are presented in Table~\ref{tab:eval_baseline_small} and Figure~\ref{fig:eval_im_summary}. 
\MethodNameShort{} once again outperforms all the baselines across all queue sizes. The percentage of improvement in performance over the realistic baselines is 22\%-76\%, while the improvement over the unrealistic baselines is 13\%-64\%. The results clearly show that our hierarchical approach to the modeling of large queues is very effective in enabling DRL-based solutions to scale to larges queue sizes.

\begin{figure}[h]
    \centering
    \includegraphics[scale = 0.5]{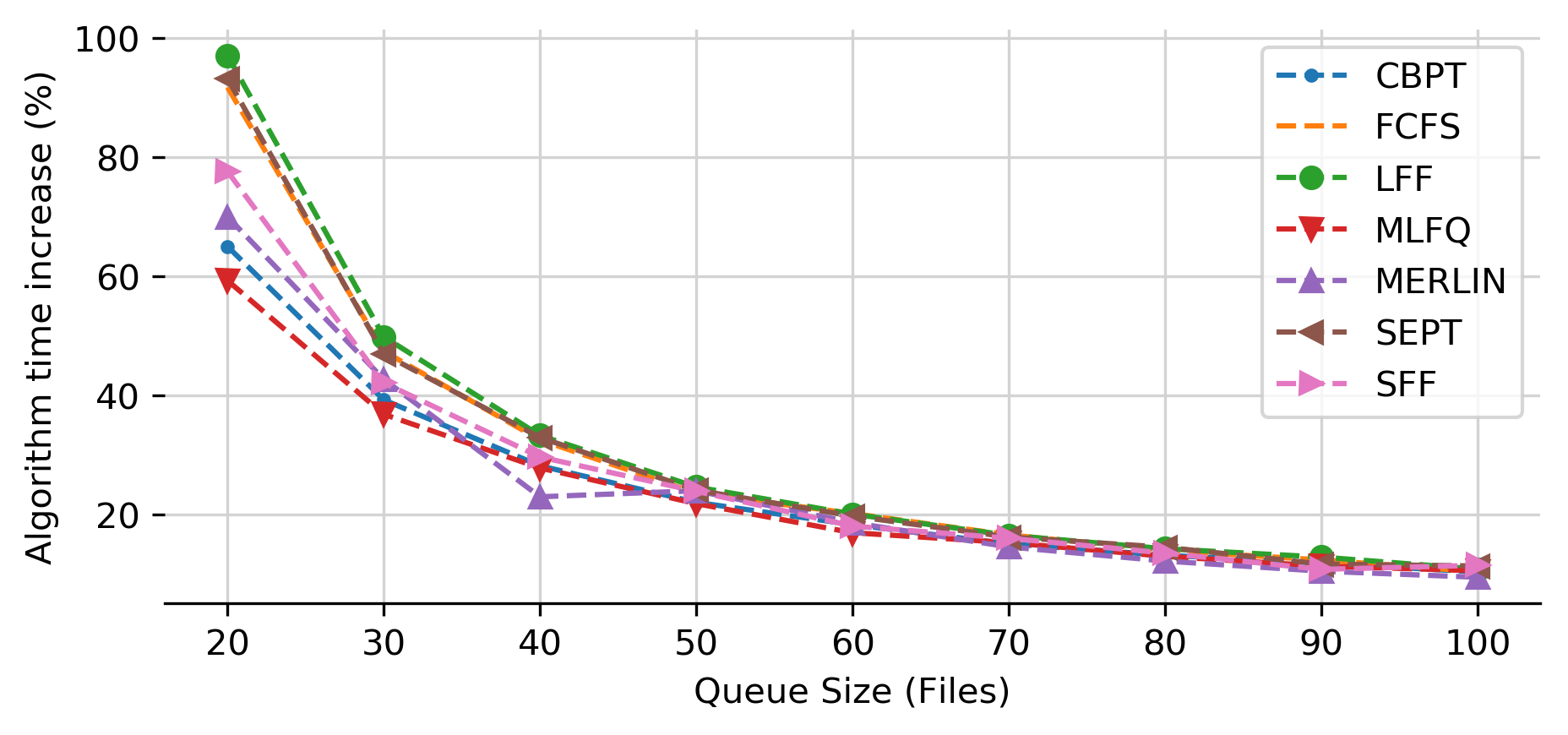}
    \caption{The percentage of the increase in average file processing time as a function of the queue size (with respect to the algorithm's performance on $|Q|=10$).}
    \label{fig:eval_baseline_im}
\end{figure}

Additionally, we were interested in determining whether \MethodNameShort{}'s superior performance is likely to subsists for larger queue sizes. We therefore analyzed the increase in average file processing time as a function of the queue size for each of the evaluated algorithms. The results, presented in Figure \ref{fig:eval_baseline_im}, show that for the increase in average processing time plateaus relatively quickly. The meaning is that methods that achieved better performance for smaller queue sizes are likely to maintain their relative lead in lager queue sizes (it is important to note that each algorithm in Figure \ref{fig:eval_baseline_im} is measured with respect to itself).

\subsubsection{Experiment 3: Dynamic Queues with Stochastic Arrivals}

\begin{figure*}[ht]
    \centering
    \includegraphics[scale = 0.47]{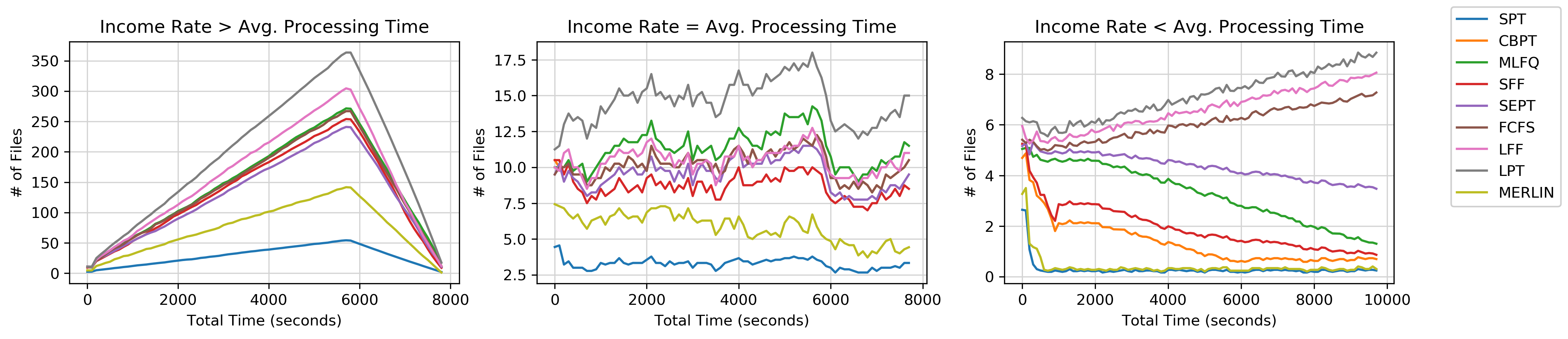}
    \caption{Queue size behavior in different entry rates.
    The graphs show three use-cases of incoming rates respectively from left to right: incoming rate $ > $ average processing time, incoming rate = average processing time, and incoming rate $ < $ average processing time. \MethodNameShort{} has showed stability and consistency in its results with respect to the other methods.}
    \label{fig:eval_dynamic_comparsion}
\end{figure*}

In most real-world scenarios, queues are \textit{dynamic}, with new items being added at various time intervals.
This is the case for call centers, manufacturing floors, and (as in our use-case) organizational firewall that filter incoming files. 
This scenario adds another level of complexity compared with previous experiments, because the scheduling algorithms need to predict the number and characteristics of the incoming files.

We evaluated three use-cases (i.e., different scenarios). 
In the first use-case, the incoming files rate is higher than the average processing time. 
This means that the backlog will grow for all approaches, and that their test will be slowing this growth. 
In the second use-case, the incoming files rate is equal to the average processing time. 
In this case we expect the size of the backlog to be stable, and the scheduling algorithms will be evaluated based on the size of the backlog they keep. 
In the third use-case the incoming files rate is lower than the average processing time, and the scheduling algorithms will be evaluated based on their ability to keep the backlog as close as possible to zero. 

The time interval for adding new files to the queue was identical for all three use-cases.
Through the analysis of our training set, we've learned that the average file processing time as $\mu=7.8$ seconds with a standard deviation of $\sigma=1.9$ seconds. 
Therefore, every 7.8 seconds, we randomly sampled a fixed number of files for each use case: for the first experiment, the number of files was $\mu-\sigma$, for the second use case, the number was $\mu$ and for the third use case, the number was $\mu+\sigma$. 
The variance in performance between the added file batches stems from the fact that the files of each batch are samples randomly and therefore their characteristics vary. 

In all use-cases, we sampled 1,000 files overall and recorded the backlog of each analyzed algorithm until the backlog was cleared. 
The results of our experiments are presented in Figure~\ref{fig:eval_dynamic_comparsion}, and they clearly show that \MethodNameShort{} significantly outperformed all baselines (except for the optimal baseline, which we use as a bound). 
In all three scenarios, the backlog kept by our approach was the smallest, often by a significant margin. 
These results once again illustrate the effectiveness of our proposed approach in general and that of our hierarchical representation in particular.

\subsubsection{Analysis: Training Larger DRL architectures}
\label{subsubsec:runningTimes}
We now demonstrate the significant efficiently that can be obtained by using our proposed hierarchical approach for analyzing larger queues. To this end, we trained two additional \MethodNameShort{} architectures, where the sizes of the input and output layers of the outer agent were enlarged so that the architecture could analyze queue sizes of 20 and 30 respectively. 

We compare the running times of the two new architectures to the original \MethodNameShort{} architecture. The results, presented in Table~\ref{tab:eval_trained_comp} show that in order to achieve comparable performance to that of the original \MethodNameShort{} ($|Q|=10$), the larger architectures need to run for significantly longer periods of time. For example, in order to reach the same final average processing time as the original \MethodNameShort{}, the $|Q|=20$ version needs to run almost seven times as long. For the $|Q|=30$ we were not even to obtain full convergence on our hardware and had to terminate the experiment.

\begin{table}[h]
\centering
\small
\caption{A Comparison of the training time (hr) and the Avg. completion time for a file in the queue (sec) between the original \MethodNameShort{} architecture ($Q=10$) and two larger architectures trained on $Q=20/30$ respectively.}
\begin{tabular}{c|c|cc|cc}
\toprule
 & \# & \multicolumn{2}{c|}{Training Time} & \multicolumn{2}{c}{Avg. Completion} \\
  Queue&  of & \multicolumn{2}{c|}{(Hours)} & \multicolumn{2}{c}{Time (Seconds)} \\
 \cline{3-4} \cline{5-6}
 Size &  Epochs& \MethodNameShort{} & \MethodNameShort{} &\MethodNameShort{} & \MethodNameShort{} \\
  &  & $Q = 10$ & $Q = 20/30$ & $Q=10$& $Q=20/30$ \\
\midrule
\multirow{5}{*}{20} & 10 & 5 & 19 & 34.46 & 39.62 \\
 & 20 & 12 & 42 & 31.48 & 36.64 \\
 & 34 & 22 & 73 & 28.74 & 30.24 \\
 & 40 & - & 102 & - & 30.04 \\
 & 58 & - & 141 & - & 29.86 \\
\midrule
\multirow{3}{*}{30} & 10 & 5 & 24 & 49.17 & 65.19 \\
 & 20 & 12 & 51 & 44.91 & 61.47 \\
 & 34 & 22 & 95 & 41.01 & 53.88 \\
\bottomrule

\end{tabular}
\label{tab:eval_trained_comp}
\end{table}

\section{\label{sec:related_work}RELATED WORK}

The field of MOTS-related solutions is diverse both in its techniques and the domains in which it is applies. For example, in the field of operations management, the authors of ~\cite{koksalan2003using} developed a heuristic approach based on a genetic algorithm for the bi-objective scheduling on a single machine problem of minimizing flow time (i.e., total processing time) and number of tardy (i.e., measure of a delay in execution) jobs. In the semiconductor manufacturing domain, the work presented in~\cite{gupta2005single} addressed the problem of scheduling $n$ independent jobs on a single testing machine with due dates and sequence-dependent setup times.

RL has been proven as an efficient method for scheduling. 
In the computing memory control domain, RL was used for resource allocation: in their paper \cite{p30}, the authors introduced a memory control-based scheduling algorithm using RL to fully utilize dynamic random-access memory (DRAM) bandwidth. This goal was achieved by observing the system state and estimating the long-term performance impact of each predefined possible actions of the processor.
In the cloud computing domain~\cite{mao2016resource}, the authors presented a RL-based scheduling method in a multi resource cluster environment uses the preliminary data of job duration and resource requirements per job.
In the application domain~\cite{p113}, the authors presented an automated repair system that uses RL algorithm for scheduling and allocating repairs based on previously defined constraints.
Our review of the related works indicates that there are no implementations of RL where scheduling was conducted under uncertainty (i.e., no knowledge on the expected processing time of each job).

RL for multi-objective scheduling was also introduced in several domains. 
In manufacturing systems domain, \cite{aissani2009dynamic} presented a RL-based approach for efficient machine maintaining tasks.
Their objective was to assign maintenance tasks between different resources while minimizing resource down-time due to maintenance. 
Another implementation, conducted in the grid computing resource allocation domain~\cite{perez2010multi}, proposed the use of RL for minimizing the waiting time tasks to computational resources when the resources are under the use of other tasks.
To the best of our knowledge, there were no cases of using RL for multi objective scheduling that included both resource allocation optimization and task scheduling of a queue.

When referring the problem of changes in the state-action space, there are hardly any solutions since RL relies on a fixed state-action space.
In their paper, \cite{heffetz2019deepline} showed an approach for dynamic actions modeling that enables a RL agent to model a varying number of actions using a fixed-size representation. 
They devise a hierarchical representation of the actions space, where each level of the hierarchy is split into equal sized clusters of the actions. 
The agent iterates over the clusters of each level, selecting one action per cluster. 
The chosen actions are passed to the next level of the hierarchy, which is then also clustered. 
The process is repeated until it reaches a hierarchy level in which there are $n$ actions at most, among these actions one is chosen. 
We adopt this method and fit it to our specific needs.

\noindent \textbf{Hierarchical reinforcement learning.} The field of hierarchical reinforcement learning (HRL) is a computational approach intended to address issues of RL such as large action/state space and generalization of complex environments by learning a policy made up of multiple layers, each of which is responsible for control at a different level of temporal abstraction.

HRL has been proposed in several forms. The feudal learning~\cite{dayan1993feudal} approach takes advantage of two notions: the managerial hierarchy observes the environment at different resolutions and the communication is made between managers and "workers" through goals - a reward is given for reaching them. In~\cite{bacon2017option}, the authors showed another two levels-based approach, where the bottom level is responsible of output actions given a sub-policy, and a top level which outputs sub-policy given a policy-over-options.
Another approach presented in~\cite{dietterich2000hierarchical}, obtains the task of hierarchy by decomposing the Q value of state-action pair into the sum of two components - the the total expected reward received when executing the action in the current state and the total reward expected from the performance of the parent-task.

It is very important to note that all the approaches mentioned above, and to the best of our knowledge \textit{all} HRL methods, do not address the challenges associated with MOTS problems, and particularly those which have conflicting goals.

\section{CONCLUSIONS}
We presents \MethodNameShort{}, a two-stage DRL-based scheduling approach specifically designed to tackle the challenges of multi-objective optimization and high levels of uncertainty. 
Through extensive evaluation, we show that \MethodNameShort{} outperforms several commonly-used baselines by a wide margin. 
Additionally, we present a novel hierarchical approach for applying DRL on dynamic queues.
The ability to train the DRL-agent over smaller queue sizes enables the use of smaller architectures and leads to shorter convergence times.
For future work, we plan to further explore the application of \MethodNameShort{} in additional domains.



\bibliographystyle{ACM-Reference-Format}
\bibliography{sections/references}

\end{document}